\documentclass[a4paper]{article}

\usepackage{INTERSPEECH2016}

\usepackage{graphicx}
\usepackage{amssymb,amsmath,bm}
\usepackage{textcomp}
\usepackage{CJK}

\usepackage{times}
\usepackage{latexsym}
\usepackage{graphicx}
\usepackage{multirow,paralist,color}

\ninept

\title{Dialogue Session Segmentation by Embedding-Enhanced TextTiling}

\makeatletter
\def\name#1{\gdef\@name{#1\\}}
\makeatother \name{\em Yiping Song,$^{1}$ Lili Mou,$^{1,2}$
Rui Yan,$^{1,3,4,*}$ Li Yi,$^4$ Zinan Zhu,$^4$ Xiaohua Hu,$^4$ Ming Zhang$^1$}
\address{$^1$Peking University, Beijing, China \\
  $^2$Key Laboratory of High Confidence Software Technologies (Peking University), MoE, China \\
  $^3$Natural Language Processing Department, Baidu Inc., Beijing, China\ \  $^*$Corresponding author\\
  $^4$School of Computer, Central China Normal University, Wuhan, Hubei, China\\
  {\small \tt songyiping@pku.edu.cn, doublepower.mou@gmail.com}\\
  {\small \tt \{yanrui,yili,zzn,huxiaohua\}@mail.ccnu.edu.cn}\\
  {\color{blue}In \textit{Proc. INTERSPEECH}, pages 2706--2710, 2016.}
}

\begin{document}

  \maketitle
  \vspace{-1cm}
  \begin{abstract}
In human-computer conversation systems, the context of a user-issued utterance is particularly important because it provides useful background information of the conversation. However, it is unwise to track all previous utterances in the current session as not all of them are equally important. In this paper, we address the problem of session segmentation. We propose an embedding-enhanced TextTiling approach, inspired by the observation that conversation utterances are highly noisy, and that word embeddings provide a robust way of capturing semantics. Experimental results show that our approach achieves better performance than the TextTiling, MMD approaches.
  \end{abstract}

\section{Introduction}
Human-computer dialog/conversation\footnote{A full dialog system typically involves speech recognition, text understanding, and speech synthesis. In this paper, we focus on the text understanding stage. However, our approach is directly applicable to dialogue systems with acoustic interaction, provided that the spoken language is converted to texts by automatic speech recognition (ASR) \cite{asu2,liu2015}, or even manually text-transcribed for research purposes like \cite{manual,manual2}.} is one of the most challenging problems in artificial intelligence. Given a user-issued utterance (called a \textit{query} in this paper), the computer needs to provide a \text{reply} to the query. In early years, researchers have developed various domain-oriented dialogue systems, which are typically based on rules or templates \cite{rule,rule3,rule2}. Recently, open-domain conversation systems have attracted more and more attention in both academia and industry (e.g., XiaoBing from Microsoft and DuMi from Baidu). Due to high diversity, we can hardly design rules or templates in the open domain. Researchers have proposed information retrieval methods \cite{otterbacher2005} and modern generative neural networks \cite{responding,hseq2seq} to either search for a reply from a large conversation corpus or generate a new sentence as the reply.

\begin{figure}[!t]
\vspace{-.8cm}

\centering
\includegraphics[width=.29\textwidth]{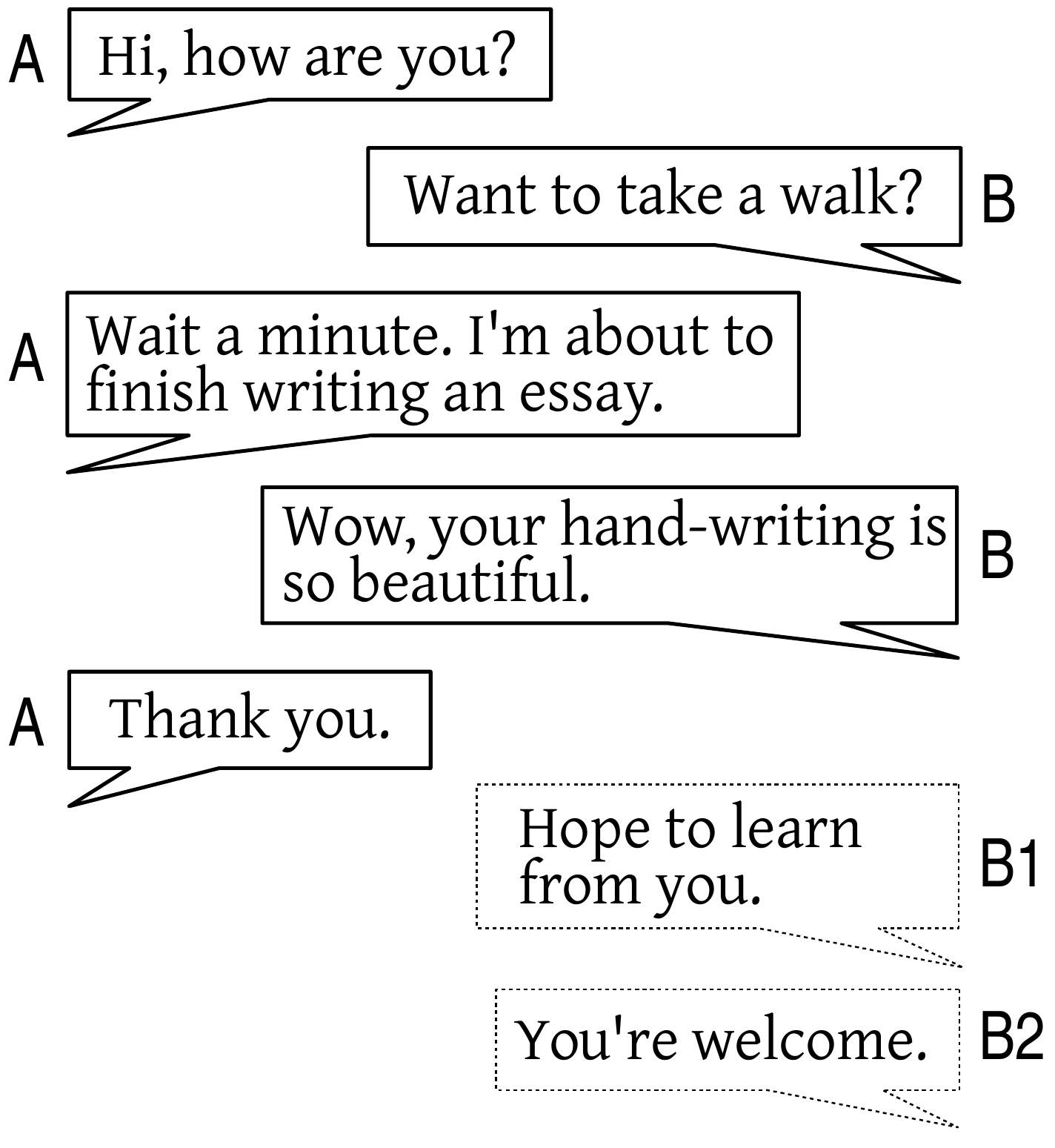}
\vspace{-.4cm}
\caption{An example of multiple-turn dialogues.}
\vspace{-.4cm}
\label{fig:example}
\end{figure}

In open-domain conversations, context information (one or a few previous utterances) is particularly important to language understanding \cite{liu2015,hseq2seq,bhargava2013,context:naacl}. As dialogue sentences are usually casual and short, a single utterance (e.g., ``Thank you.'' in Figure~\ref{fig:example}) does not convey much meaning, but its previous utterance
(``\dots writing an essay'') provides useful background information of the conversation. Using such context will certainly benefit the conversation system.

However, tracking all previous utterances as the context is unwise. First, commercial chat-bots usually place high demands on efficiency. In a retrieval-based system, for example, performing a standard process of candidate retrieval and re-ranking for each previous utterance may well exceed the time limit (which is very short, e.g., 500ms).
Second, we observe that not all sentences in the current conversation session are equally important. The sentence ``Want to take a walk?'' is irrelevant to the current context, and should not be considered when the computer synthesizes the reply. Therefore, it raises the question of \textit{session segmentation} in conversation systems.

Document segmentation for general-purpose corpora has been widely studied in NLP. For example, Hearst~\cite{hearst1997} proposes the TextTiling approach; she measures the similarity of neighboring sentences based on bag-of-words features, and performs segmentation by thresholding.  However, such approaches are not tailored to the dialogue genre and may not be suitable for conversation session segmentation.

In this paper, we address the problem of session segmentation for open-domain conversations. We leverage the classic TextTiling approach, but enhance it with modern embedding-based similarity measures. Compared with traditional bag-of-words features, embeddings map discrete words to real-valued vectors, capturing underlying meanings in a continuous vector space; hence, it is more robust for noisy conversation corpora. Further, we propose a tailored method for word embedding learning. In traditional word embedding learning, the interaction between two words in a query and a reply is weaker than that within an utterance. We propose to combine a query and its corresponding reply as a ``virtual sentence,'' so that it provides a better way of modeling utterances between two agents.

\section{Related Work}

\subsection{Dialogue Systems and Context Modeling}

Human-computer dialogue systems can be roughly divided into several categories.
Template- and rule-based systems are mainly designed for certain domains \cite{rule,rule3,watanabe2014}.
Although manually engineered templates can also be applied in the open domain like \cite{7forms}, but their generated sentences are subject to 7 predefined forms, and hence are highly restricted.
Retrieval methods search for a candidate reply from a large conversation corpus given a user-issued utterance as a query \cite{otterbacher2005}. Generative methods can synthesize new replies by statistical machine translation \cite{haas2015,tavernier2008} or neural networks \cite{responding}.

% context-free

The above studies do not consider context information in reply retrieval or generation. However, recent research shows that previous utterances in a conversation session are important because they capture rich background information. Sordoni et al.~\cite{context:naacl} summarize a single previous sentence as  bag-of-words features, which are fed to a recurrent neural network for reply generation. Serban et al.~\cite{serban2015} design an attention-based neural network over all previous conversation turns/rounds, but this could be inefficient if a session lasts long in real commercial applications. By contrast, our paper addresses the problem of session segmentation so as to retain near, relevant context utterances and to eliminate far, irrelevant ones.

A similar (but different) research problem is topic tracking in conversations, e.g., \cite{kim2014,celikyilmaz2011,lagus2002,crook2015}.
In these approaches, the goal is typically a classification problem with a few pre-defined conversation states/topics, and hence it can hardly be generalized to general-purpose session segmentation.

\subsection{Text Segmentation}

An early and classic work on text segmentation is TextTiling, proposed in \cite{hearst1997}.
The idea is to measure the similarity between two successive sentences with smoothing techniques; then segmentation is accomplished by thresholding of the depth of a ``valley.'' In the original form of TextTiling, the cosine of term frequency features is used as the similarity measure.
%Latent semantic analysis (LSA) and latent Dirichlet allocation (LDA) are applied as the feature of a sentence \cite{deerwester1990}. 
Joty et al.~\cite{joty2013} apply divisive clustering instead of thresholding for segmentation.  Malioutov et al.~\cite{malioutov2006} formalize segmentation as a graph-partitioning problem and propose a minimum cut model based on \textit{tf}$\cdot$\textit{idf} features to segment lectures. Ye et al.~\cite{ye2007} minimize between-segment similarity while maximizing within-segment similarity.
However, the above complicated approaches are known as \textit{global} methods: when we perform segmentation between two successive sentences, future context information is needed. Therefore, they are inapplicable to real-time chat-bots, where conversation utterances can be viewed as streaming data.

In our study, we prefer the simple yet effective TextTiling approach for open-domain dialogue session segmentation, but enhance it with modern advances of word embeddings, which are robust in capturing semantics of words. We propose a tailored algorithm for word embedding learning by combining a query and context as a ``virtual document''; we also propose several heuristics for similarity measuring.

%
% A:  Hi, how are you?
% B:  Want to take a walk?
% A:  Wait a minute. I'm about to finish writing an essay.
% B:  Wow, your hand-writing is so beautiful.
% A:  Thank you.
% B1? Hope to learn from you.
% B2: You're welcome.

\section{Session Segmentation Methodology}
\subsection{TextTiling}
We apply a TextTiling-like algorithm for session segmentation. The original TextTiling is proposed by Hearst~\cite{hearst1997}. The main idea is to measure the similarity of each adjacent sentence pair; then ``valleys'' of similarities are detected for segmentation.

Concretely, the ``depth of the valley'' is defined by the similarity differences between the peak point in each side and the current position. We may obtain some statistics of depth scores like the mean $\mu$ and standard deviation $\sigma$, and perform segmentation by a cutoff threshold.
\begin{figure}[!t]
\centering
\includegraphics[width=.37\textwidth]{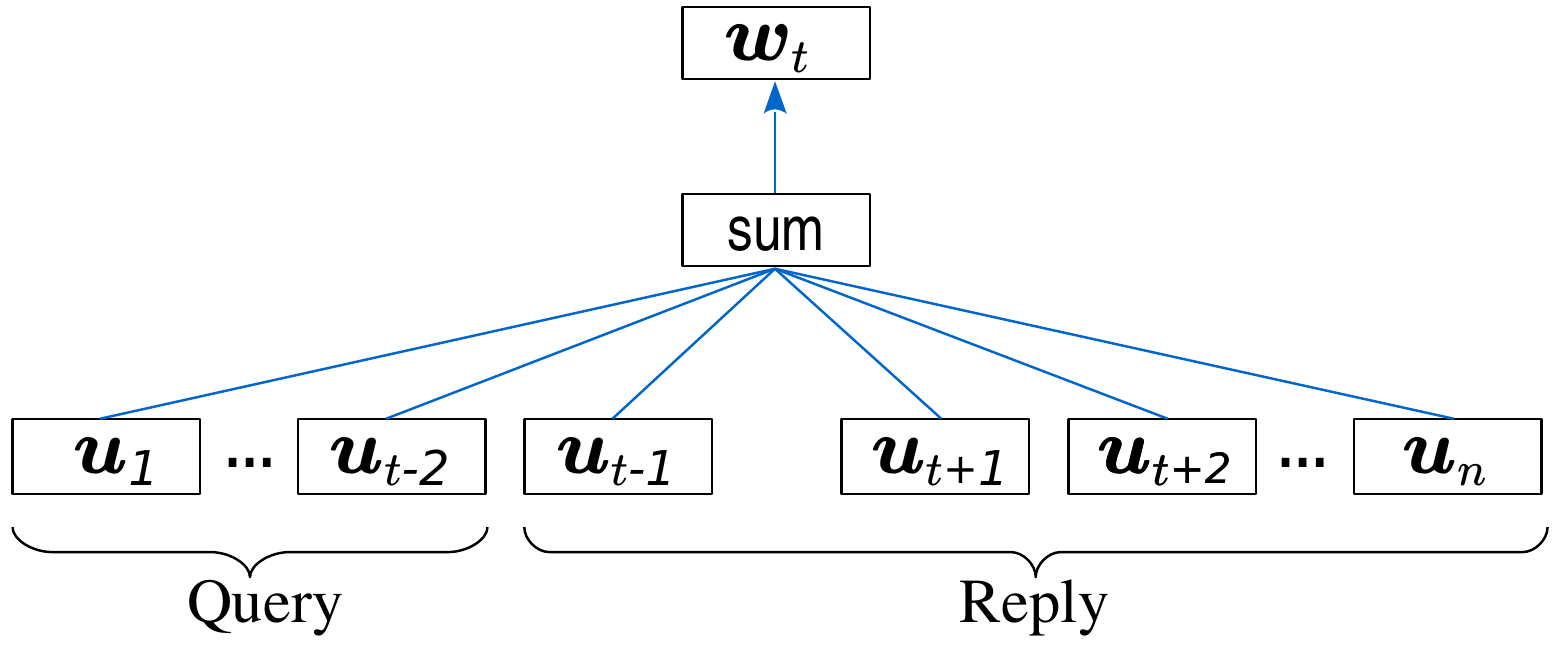}
\caption{Word embedding learning by the continuous bag-of-words model with virtual sentences (the concatenation of a query and its reply). $w_t$ is a word in the virtual sentence, either appearing in the query or the reply; the summed embeddings of remaining words are context.}
\end{figure}
\begin{equation}\textit{cutoff}(\alpha) = \mu + \alpha \cdot \sigma \label{eqn}\end{equation}
where $\alpha$ is a hyperparameter adjusting the number of segmentation boundaries; $\mu$ and $\sigma$ are the average and standard deviation of depth scores, respectively.

In the scenario of human-computer conversations, we compute the depth solely by the similarity difference between its left peak (previous context) and the current position. This is because we cannot obtain future utterances during online conversation.

Although bag-of-words features work well in the original TextTiling algorithm for general text segmentation, it is not suitable for dialogue segmentation. As argued by Hearst~\cite{hearst1997}, text overlap (repetition) between neighboring sentences is a strong hint of semantic coherence, which can be well captured by term frequency or \textit{tf}$\cdot$\textit{idf} variants. However, in human-computer conversations, sentences are usually short, noisy, highly diversified, and probably incomplete,  which requires a more robust way of similarity measuring.  Therefore, we enhance TextTiling with modern word embedding techniques, as will be discussed in the next part.

\subsection{Learning Word Embeddings}

Word embeddings are distributed, real-valued vector representations of discrete words \cite{mikolov2013a,mikolov2013b}.
Compared with one-hot representation, word embeddings are low-dimensional and dense, measuring word meanings in a continuous vector space. Studies show that the offset of two words' embeddings represents a certain relation, e.g., ``\texttt{man}'' $-$ ``\texttt{woman}'' $\approx$ ``\texttt{king}'' $-$ ``\texttt{queen}''~\cite{mikolov2013a}. Hence, it is suitable to use word embeddings to model short and noisy conversation utterances.

To train the embeddings, we adopt the \texttt{word2vec} approach. The idea is to map a word $w$ and its context $c$ to vectors ($\bm w$ and $\bm c$). Then we estimate the probability of a word by
\begin{equation}
p(w|c)=\dfrac{\exp(\bm w^\top\bm c)}{\sum_{\bm w'}\exp(\bm w'^\top\bm c)}\label{eqn:w}
\end{equation}
The goal of word embedding learning is to maximize the average probability of all words (suppose we have $T$ running words):
\begin{equation}
\dfrac1T\sum_{t=1}^T\log p(w_t|c_t)
\end{equation}
We used hierarchical softmax to approximate the probability.

To model the context, we further adopt the continuous bag-of-words (CBOW) method. The context\footnote{Here, the context of a word roughly refers to its previous and future words. Please do not be confused with the context of an utterance.} is defined by the sum of neighboring words' (input) vectors in a fixed-size window ($t-\tau$ to $t+\tau$) within a sentence:
\begin{equation}
\bm c_t=\sum_{t-\tau \le i\le t+\tau \atop i\ne t}\bm u_i\label{eqn:c}
\end{equation}
Notice that the context vector $\bm u$ in Equation~(\ref{eqn:c}) and the output vector $\bm w$ in Equation~(\ref{eqn:w}) are different as suggested in \cite{mikolov2013a,mikolov2013b}, but the details are beyond the scope of our paper.

\medskip
\noindent\textbf{Virtual Sentences}

In a conversation corpus, successive sentences have a stronger interaction than general texts. For example, in Figure~\ref{fig:example}, the words \textit{thank} and \textit{welcome} are strongly correlated, but they hardly appear in the a sentence and thus a same window. Therefore, traditional within-sentence CBOW may not capture the interaction between a query and its corresponding reply.

In this paper, we propose the concept of virtual sentences to learn word embeddings for conversation data. We concatenate a query $q$ and its reply $r$ as a \textit{virtual sentence} $q\oplus r$. We also use all words (other than the current one) in the virtual sentence as context (Figure 2). Formally, the context $c_t$ of the word $w_t$ is given by
\begin{equation}
\bm c_t= \sum_{i\in q\oplus r\atop i\ne t}\bm u_i
\end{equation}
In this way, related words across two successive utterances from different agents can have interaction during word embedding learning. As will be shown in Subsection~\ref{ss:results}, virtual sentences yield a higher performance for dialogue segmentation.

\subsection{Measuring Similarity}

In this part, we introduce several heuristics of similarity measuring based on word embeddings.
Notice that, we do not leverage supervised learning (e.g., full neural networks for sentence paring \cite{pair1,pair2}) to measure similarity, because it is costly to obtain labeled data of high quality.

The simplest approach, perhaps, is to sum over all word embeddings in an utterance as sentence-level features $\bm s$. This heuristic is essentially the sum pooling method widely used in neural networks \cite{lenet5,tbcnn,pair1}.
The cosine measure is used as the similarity score between two utterances $S_1$ and $S_2$. Let $\bm s_1$ and $\bm s_2$ be their sentence vectors; then we have
\begin{equation}
\text{sim}(S_1,S_2)=\cos(\bm s_1,\bm s_2)\equiv\dfrac{\bm s_1^\top\bm s_2}{\|\bm s_1\|\cdot\|\bm s_2\|}
\end{equation}
where $\|\cdot\|$ is the $\ell_2$-norm of a vector.

To enhance the interaction between two successive sentences, we propose a more complicated heuristic as follows. Let $w_i$ and $v_j$ be a word in $s_1$ and $s_2$, respectively. (Embeddings are denoted as bold alphabets.) Suppose further that $n_1$ and $n_2$ are the numbers of words in $S_1$ and $S_2$. The similarity is given by
\begin{equation}
\text{sim}(S_1,S_2)= \frac{1}{n_1}\sum_{i=1}^{n_1}\ {\max\nolimits_{j=0}^{n_2}\{\cos(\bm w_i,\bm v_j)\}}\label{eqn:heuristics}\end{equation}

For each word $w_i$ in $s_1$, our intuition is to find the most related word in $s_2$, given by the $\max\{\cdot\}$ part; their relatedness is also defined by the cosine measure. Then the sentence-level similarity is obtained by the average similarity score of words in $s_1$.
This method is denoted as \textbf{\emph{heuristic-max}}.

Alternatively, we may substitute the $\max$ operator in Equation~(\ref{eqn:heuristics}) with $\operatorname{avg}$, resulting in the \textbf{\emph{heuristic-avg}} variant, which is equivalent to the average of word-by-word cosine similarity. However, as shown in Subsection~\ref{ss:results}, intensive similarity averaging has a ``blurring'' effect and will lead to significant performance degradation. This also shows that our proposed \textit{heuristic-max} does capture useful interaction between two successive utterances in a dialogue.

\section{Experiments}
In this section, we evaluate our embedding-enhanced TextTiling method as well as the effect of session segmentation. In Subsection~\ref{ss:dataset}, we describe the datasets used in our experiments. Subsection~\ref{ss:results} presents the segmentation accuracy of our method and baselines. In Subsection~\ref{ss:external}, we show that, with our session segmentation, we can improve the performance of a retrieval-based conversation system.

\subsection{Dataset}
\label{ss:dataset}
% 200 100, 100
\begin{table}[!t]
\centering
\begin{tabular}{|l|c|c|c|}
\hline
\!\!Method &\!\! P  \!\! & \!\!R \!\!& \!\!F\!\! \\ \hline
\hline
\!\!Random                                                 &\!\! 36.9 \!\!& \!\!32.0\!\! & \!\!34.2\!\!\\
\!\!MMD                                                    &\!\! 27.0 \!\!& \!\!23.0\!\! & \!\!25.0\!\!\\
\!\!TextTiling with \textit{tf}$\cdot$\textit{idf}         &\!\! 66.5 \!\!&\!\! 44.8 \!\!& \!\!53.5\!\!\\
\hline
\hline
\!\!$+$ embeddings trained by virtual                    & \!\!\multirow{2}{*}{\textbf{71.7}}\!\! & \!\!\multirow{2}{*}{\textbf{78.0}}\!\! & \!\!\multirow{2}{*}{\textbf{74.7}}\!\!\\
\!\!sentences, and \textit{heuristic-max} similarity\!\!   & & & \\
\hline
\end{tabular}
\caption{Dialogue session segmentation performance in terms of precision (P), recall (R) and $F$-measure (F). Results are in percentage.}\label{tab:performance}
\end{table}
To evaluate the session segmentation method, we used a real-world chatting corpus from DuMi,\footnote{http://xiaodu.baidu.com} a state-of-the-practice open-domain conversation system in Chinese. We sampled 200 sessions as our experimental corpus.
Session segmentation was manually annotated before experiments, serving as the ground truth.
The 200 sessions were randomly split by 1:1 for validation and testing. Notice that, our method does not require labeled training samples; massive data with labels of high quality are quite expensive to obtain.

We also leveraged an unlabeled massive dataset of conversation utterances to train our word embeddings with ``virtual sentences.'' The dataset was crawled from the Douban forum,\footnote{http://www.douban.com} containing 3 million utterances and approximately 150,000 unique words (Chinese terms).

\begin{table*}[!t]
\centering
\begin{tabular}{|l||ccc|ccc|ccc|}
\hline
\multirow{2}{*}{Embeddings}   & \multicolumn{3}{c|}{{sum pooling}}  &  \multicolumn{3}{c|}{\textit{heuristic-max}}  &  \multicolumn{3}{c|}{\textit{heuristic-avg}}  \\
 \cline{2-10}
 & P & R & F   & P & R & F     & P & R &  F\\
\hline
\hline
Virtual-sentence  &    71.6 & 78.4 & \textbf{\underline{74.8}}   &    71.7 & 78.0 & \underline{74.7}       &    64.5 & 67.8 & \underline{66.1} \\
Within-sentence   &    66.1 & 73.1 & 69.4   &    70.0 & 74.3 & \textbf{72.1}       &    62.7 & 67.4 & 65.0 \\
Window context &    65.6 & 69.5 & 67.5   &    70.7 & 76.8 & \textbf{73.6}       &    62.2 & 65.4 & 63.8 \\
\hline
\end{tabular}
\caption{Analysis of word embedding strategies and similarity heuristics. Bold numbers are the highest value in each row; underlined ones are the highest in each column.}
\label{tab:analysis}
\end{table*}

\subsection{Segmentation Performance}
\label{ss:results}
We compared our full method (TextTiling with \textit{heuristic-max} based on embeddings trained by virtual sentences) with several baselines:
\begin{compactitem}
\item \textit{Random.} We randomly segmented conversation sessions. In this baseline, we were equipped with the prior probability of segmentation.
\item \textit{MMD.} We applied the MinMax-Dotplotting (MMD) approach proposed by Ye et al.~\cite{ye2007}.
We ran the executable program provided by the authors.
\item \textit{TextTiling w/ tf$\cdot$idf features.} We implemented TextTiling ourselves according to \cite{hearst1997}.
\end{compactitem}

\begin{table}[!t]
\centering
\begin{tabular}{|c|cc|}
\hline
Method & p@1 & nDCG\\
\hline\hline
Fixed context & 0.484 & 0.699\\
Session segmentation & 0.521 & 0.737\\
\hline
\end{tabular}
\caption{A retrieval dialogue system with fixed context (2 previous utterances) and the proposed sentence segmentation (virtual sentences with \textit{heuristic-max}).}
\label{tab:external}

\end{table}

We tuned the hyperparameter $\alpha$ in Equation~(\ref{eq})on the validation set to make the number of segmentation close to that of manual annotation, and reported precision, recall, and the F-score on the test set in Table~\ref{tab:performance}. As seen, our approach significantly outperforms baselines by a large margin in terms of both precision and recall. Besides, we can see that MMD obtains low performance, which is mainly because the approach cannot be easily adapted to other datasets like short sentences of conversation utterances. In summary, we achieve an $F$-score higher than baseline methods by more than 20\%, showing the effectiveness of enhancing TextTiling with modern word embeddings.

We further conducted in-depth analysis of different strategies of training word-embeddings and matching heuristics in Table~\ref{tab:analysis}. For word embeddings, we trained them on the 3M-sentence dataset with three strategies: (1) virtual-sentence context proposed in our paper; (2)
within-sentence context, where all words (except the current one) within a sentence  (either a query or reply) are regarded as the context; (3) window-based context, which is the original form of~\cite{mikolov2013a}: the context is the words in a window (previous 2 words and future 2 words in the sentence).
We observe that our virtual-sentence strategy consistently outperforms the other two in all three matching heuristics. The results suggest that combining a query and a reply does provide more information in learning dialogue-specific word embeddings.

Regarding matching heuristics, we find that in the second and third strategies of training word embeddings, the complicated \textit{heuristic-max} method yields higher $F$-scores than simple sum pooling by 2--3\%.
However, for the virtual-sentence strategy, \textit{heuristic-max} is slightly worse than the sum pooling. (The degradation is only 0.1\% and not significant.)
This is probably because both \textit{heuristic-max} and virtual sentences emphasize the rich interaction between a query and its corresponding reply; combining them does not result in further gain.

We also notice that \textit{heuristic-avg} is worse than other similarity measures. As this method is mathematically equivalent to the average of word-by-word similarity, it may have an undesirable blurring effect.

To sum up, our experiments show that both the proposed embedding learning approach and the similarity heuristic are effective for session segmentation. The embedding-enhanced TextTiling approach largely outperforms baselines.

\subsection{Session Segmentation in Dialogue Systems}
\label{ss:external}
We conducted an external experiment to show the effect of session segmentation in dialogue systems.
We integrated the segmentation mechanism into a state-of-the-practice retrieval-based system and evaluated the results by manual annotation, similar to our previous work \cite{pair1,ijcai,bf}.

Concretely, we compared our session segmentation with fixed-length context, used in \cite{context:naacl}. That is to say, the competing method always regards two previous utterances as context. We hired three workers to annotate the results with three integer scores ($0$--$2$ points, indicating bad, borderline, and good replies, respectively.) We sampled 30 queries from the test set of 100 sessions. For each query, we retrieved 10 candidates and computed p@1\footnote{Out of a rigorous criterion, we regard a ``correct'' reply if the score is $2$, and ``incorrect'' if the score is $0$ or $1$.} and nDCG scores \cite{jarvelin2002} (averaged over three annotators).  Provided with previous utterances as context, each worker had up to 1000 sentences to read during annotation.

Table~\ref{tab:external} presents the results of the dialogue system with session segmentation.
As demonstrated, our method outperforms the simple fixed-context approach in terms of both metrics.
We computed the inner-annotator agreement: std $=$ 0.309; 3-discrete-class Fleiss' kappa score $=$ 0.411, indicating moderate agreement \cite{fleiss}.

\medskip
\noindent\textbf{Case Study.} We present a case study on our website: {https://sites.google.com/site/sessionsegmentation/}. From the case study, we see that the proposed approach is able to segment the dialogue session appropriately, so as to better utilize background information from a conversation session.
\section{Conclusion}
\label{sect:pdf}
In this paper, we addressed the problem of session segmentation for open-domain dialogue systems.
We proposed an embedding-enhanced TextTiling approach, where we trained embeddings with the novel notion of virtual sentences; we also proposed several heuristics for similarity measure. Experimental results show that both our embedding learning and similarity measuring are effective in session segmentation, and that with our approach, we can improve the performance of a retrieval-based dialogue system.

\section{Acknowledgments}
We thank anonymous reviewers for useful comments and Jingbo Zhu for sharing the MMD executable program. This paper is partially supported by the National Natural Science Foundation of China (NSFC Grant Nos.~61272343 and 61472006), the Doctoral Program of Higher Education of China (Grant No.~20130001110032), and the National Basic Research Program (973 Program No.~2014CB340405).

\eightpt
\bibliographystyle{IEEEtran}

\bibliography{mybib}
\end{document}